\documentclass[runningheads]{llncs}

\usepackage{eccv}

\usepackage{eccvabbrv}

\usepackage{graphicx}
\usepackage{booktabs}
\usepackage{tabu}

\usepackage[accsupp]{axessibility}  

\usepackage{multirow}
\usepackage{floatrow}
\usepackage[misc,geometry]{ifsym}
\newfloatcommand{capbtabbox}{table}[][\FBwidth]

\usepackage{hyperref}

\usepackage{orcidlink}

\begin{document}

\title{UniM$^2$AE: Multi-modal Masked Autoencoders with Unified 3D Representation for 3D Perception in Autonomous Driving} 

\titlerunning{UniM$^2$AE}

\author{Jian Zou\inst{1} \and
Tianyu Huang\inst{1}\orcidlink{0009-0002-1071-6371} \and
Guanglei Yang\inst{1}\textsuperscript{(\Letter)}\orcidlink{0000-0002-5324-3642} \and 
Zhenhua Guo\inst{2}\orcidlink{0000-0002-8201-0864} \and
Tao Luo\inst{3}\textsuperscript{(\Letter)}\orcidlink{0000-0002-3415-3676} \and
Chun-Mei Feng\inst{3}\orcidlink{0000-0002-3044-9779} \and
Wangmeng Zuo\inst{1}\orcidlink{0000-0002-3330-783X}
}

\authorrunning{J.~Zou et al.}

\institute{
Harbin Institute of Technology, China \\
\email{yangguanglei@hit.edu.cn} \and
Tianyijiaotong Technology Ltd., China \and
Institute of High Performance Computing (IHPC), Agency for Science, Technology and Research (A*STAR), Singapore \\
\email{tluo001@e.ntu.edu.sg}
}

\maketitle

\begin{abstract}
Masked Autoencoders (MAE) play a pivotal role in learning potent representations, delivering outstanding results across various 3D perception tasks essential for autonomous driving. 
In real-world driving scenarios, it's commonplace to deploy multiple sensors for comprehensive environment perception.
Despite integrating multi-modal features from these sensors can produce rich and powerful features, there is a noticeable challenge in MAE methods addressing this integration due to the substantial disparity between the different modalities.
This research delves into multi-modal Masked Autoencoders tailored for a unified representation space in autonomous driving, aiming to pioneer a more efficient fusion of two distinct modalities.
To intricately marry the semantics inherent in images with the geometric intricacies of LiDAR point clouds, we propose UniM$^2$AE.
This model stands as a potent yet straightforward, multi-modal self-supervised pre-training framework, mainly consisting of two designs. 
First, it projects the features from both modalities into a cohesive 3D volume space to intricately marry the bird's eye view (BEV) with the height dimension.
The extension allows for a precise representation of objects and reduces information loss when aligning multi-modal features.
Second, the Multi-modal 3D Interactive Module (MMIM) is invoked to facilitate the efficient inter-modal interaction during the interaction process.
Extensive experiments conducted on the nuScenes Dataset attest to the efficacy of UniM$^2$AE, indicating enhancements in 3D object detection and BEV map segmentation by 1.2\% NDS and 6.5\% mIoU, respectively. The code is available at \url{https://github.com/hollow-503/UniM2AE}.
  

\end{abstract}

\begin{figure}[!ht]
    \begin{center}
        \includegraphics[width=\linewidth]{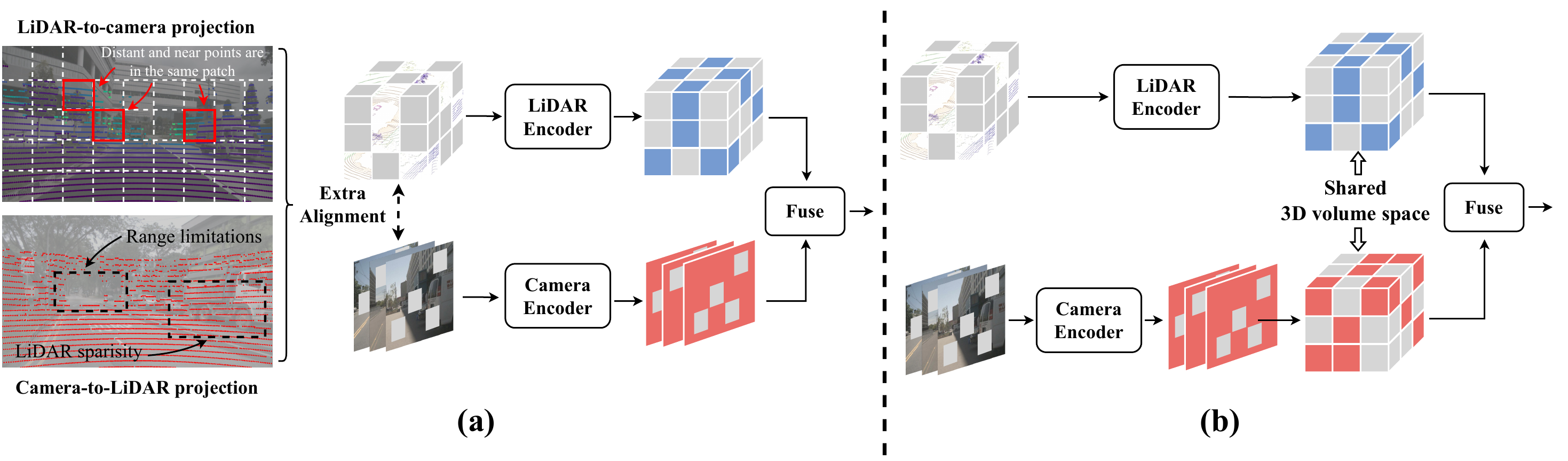}
    \end{center}
    \caption{(a) Multi-modal frameworks \cite{chen2023pimae} that align masked input before feature extraction but ignore feature characteristics from two branch. (b) UniM$^2$AE that interacts multi-modal features with unified representation. }
    \label{teaser}
    \vspace{-0.6cm}
\end{figure}

\section{Introduction}

Autonomous driving marks a transformative leap in transportation, heralding potential enhancements in safety, efficiency, and accessibility~\cite{xie2023robobev, wang2023unitr, li2022bevformer, liang2022bevfusion}. 
Fundamental to this progression is the vehicle's capability to decode its surroundings~\cite{wei2023surroundocc, zhang2023occformer, yang2023bevformer}. To tackle the intricacies of real-world contexts, integration of various sensors is imperative: cameras yield detailed visual insights, LiDAR grants exact geometric data, \etc Through this multi-sensor fusion, a comprehensive grasp of the environment is achieved~\cite{liu2023bevfusion, athar20234d}.
Nonetheless, the reliance on extensively labeled data for multi-sensor fusion poses a significant challenge due to the high costs involved.

Masked Autoencoders (MAE)~\cite{he2022masked} have shown promise in reducing reliance on labeled datasets, especially noted in 2D vision tasks.
A natural extension of this success in autonomous driving involves projecting LiDAR point clouds onto the image plane, thereby masking image patches in conjunction with their corresponding LiDAR data.
However, this approach faces a significant challenge due to geometric distortions when projecting LiDAR data onto the camera plane, as depicted in Fig.~\ref{teaser}(a).
This distortion arises due to the inherent spatial discrepancy within the same patch: point clouds that are physically distant or close may appear next to each other on the pixel coordinates, making it impractical to indiscriminately divide and mask these merged image and point cloud patches for reconstruction.
Moreover, the discrepancy in the operational range between LiDAR sensors and cameras exacerbates this challenge.
Due to the different sensor setups, LiDAR may capture point clouds beyond the camera's field of view, as seen in datasets like Waymo Open Dataset~\cite{sun2020scalability}, which provides 360-degree LiDAR views versus the camera's limited frontal and lateral coverage. As a result, a vast amount of LiDAR points are unprojectable.
Alternatively, some sensor fusion strategies\cite{sindagi2019mvx, vora2020pointpainting, wang2021pointaugmenting, yin2021multimodal} attempt to interact image and point clouds by employing the camera-to-LiDAR projection. However, this approach is hindered by the inherent sparsity of LiDAR data, resulting in a substantial loss of camera features and, consequently, a detrimental impact on effective feature fusion.

To address the above challenges, we introduce the UniM$^2$AE, depicted in Fig.~\ref{teaser}(b), as an innovative self-supervised pre-training framework designed to harmonize the integration of two distinct modalities: images and LiDAR data. 
UniM$^2$AE endeavors to establish a unified representation space that enhances the fusion of these modalities. 
By leveraging the semantic richness of images in tandem with the precise geometric details captured by LiDAR, UniM$^2$AE facilitates the generation of robust, cross-modal features.
Central to UniM$^2$AE is the innovation of a 3D volume space, achieved by extending the \textit{z}-axis of the Bird's Eye View (BEV) representation. 
This crucial expansion preserves height, enabling a more faithful representation of objects that exhibit significant variation in height. It also mitigates the information loss that is typically encountered when projecting features into or retrieving them from the representation space.
Unlike the traditional BEV representation~\cite{li2023delving} and the occupancy representation~\cite{tian2023occ3d, wei2023surroundocc}, our unified 3D representation retains sufficient details along the height dimension.
This enables accurate re-projection to the original modalities for reconstructing the multi-modal inputs.
Moreover, within this enriched 3D volume space, the information contained within each voxel can be derived either from features of its native modality or from features of other modalities. 
Each modality's decoder then utilizes this unified feature set to reconstruct its specific inputs, pushing the encoder to learn more generalized and cross-modal features that encapsulate a comprehensive understanding of the scene or objects.
This architecture not only addresses the previously outlined limitations but also sets a new benchmark for multi-modal integration, advancing the field of autonomous driving by enabling more nuanced and effective utilization of sensor data.

In the expansive and dynamic environments characteristic of autonomous driving, which feature a multitude of objects and intricate inter-instance relationships, a sophisticated approach is required for efficient feature amalgamation.
To this end, the Multi-modal 3D Interaction Module (MMIM) is employed to further refine the fusion of modalities within our unified 3D volume space, signifying a pivotal advancement in our framework's capability to facilitate efficient interaction between the dual branches of input data.
The MMIM's architecture, built upon stacked 3D deformable self-attention blocks~\cite{zhu2020deformable, xia2022vision}, enables the modeling of global context at elevated resolutions. This feature is instrumental in boosting the performance of downstream tasks, thereby addressing the complexities inherent in autonomous driving scenarios. The deformable nature of the self-attention mechanism allows for adaptive focus on the most salient features within vast scenes, ensuring that the system is attuned to the nuanced dynamics and relationships present~\cite{xia2022vision}. 
Consequently, this leads to quicker model convergence and improved pre-training efficiency, marking a significant leap forward in our endeavor to enhance the functionality of autonomous driving systems.

To sum up, our contributions can be presented as follows:
\begin{itemize}
    \item We propose UniM$^2$AE, a multi-modal self-supervised pre-training framework with unified representation in a cohesive 3D volume space. 
    The corresponding representation enables the alignment of the multi-modal features with less information loss, facilitating the reconstruction of multi-modal masked inputs.
    \item To better interact semantics and geometries retained in unified 3D volume space, we introduce a Multi-modal 3D Interaction Module (MMIM) to effectively obtain more informative and powerful features.
    \item We conduct extensive experiments on various 3D downstream tasks, where UniM$^2$AE notably promotes diverse detectors and shows competitive performance.
\end{itemize}

\section{Related Work}

\subsection{Multi-modal Representation}
Multi-modal representation has raised significant interest recently, especially in vision-language pre-training~\cite{radford2021learning,yao2021filip}. Some works~\cite{zhang2022pointclip,huang2022clip2point,yang2022deepinteraction} align point cloud data to 2D vision by depth projection. 
As for unifying 3D with other modalities, the bird's-eye view (BEV) is a widely-used representation since the transformation to BEV space retains both geometric structure and semantic density. Although many SOTA methods~\cite{liu2023bevfusion, liang2022bevfusion,chen2023focalformer3d,hu2023ea} adopt BEV as the unified representation, the lack of height information leads to a poor description of the shape and position of objects, which makes it unsuitable for MAE.
In this work, we introduce a unified representation with height dimension in 3D volume space, which captures the detailed height and position of objects.

\subsection{Masked Autoencoders}
Masked Autoencoders (MAE)~\cite{he2022masked} are a self-supervised pre-training method, with a pre-text task of predicting masked pixels. With its success, a series of 3D representation learning methods apply masked modeling to 3D data. Some works~\cite{yu2022point,pang2022masked,liu2022masked} reconstruct masked points of indoor objects. Some works~\cite{min2022voxel,tian2023geomae, xu2023mv, yang2023gd, boulch2023also} predict the masked voxels in outdoor scenes. 
Recent methods propose multi-modal MAE pre-training: Zhang~\cite{zhang2023learning} exploit 2D pre-trained knowledge to the 3D point prediction but fail to exploit the full potential of LiDAR point cloud and image datasets. Chen~\cite{chen2023pimae} attempt to tackle it in the indoor scene and conduct a 2D-3D joint prediction by the projection of points but ignore the characteristic of point clouds and images. To address these problems, we propose to predict both masked 2D pixels and masked 3D voxels in a unified representation, focusing on the autonomous driving scenario.

\begin{figure*}[!htb]
    \begin{center}
        \includegraphics[width=\linewidth]{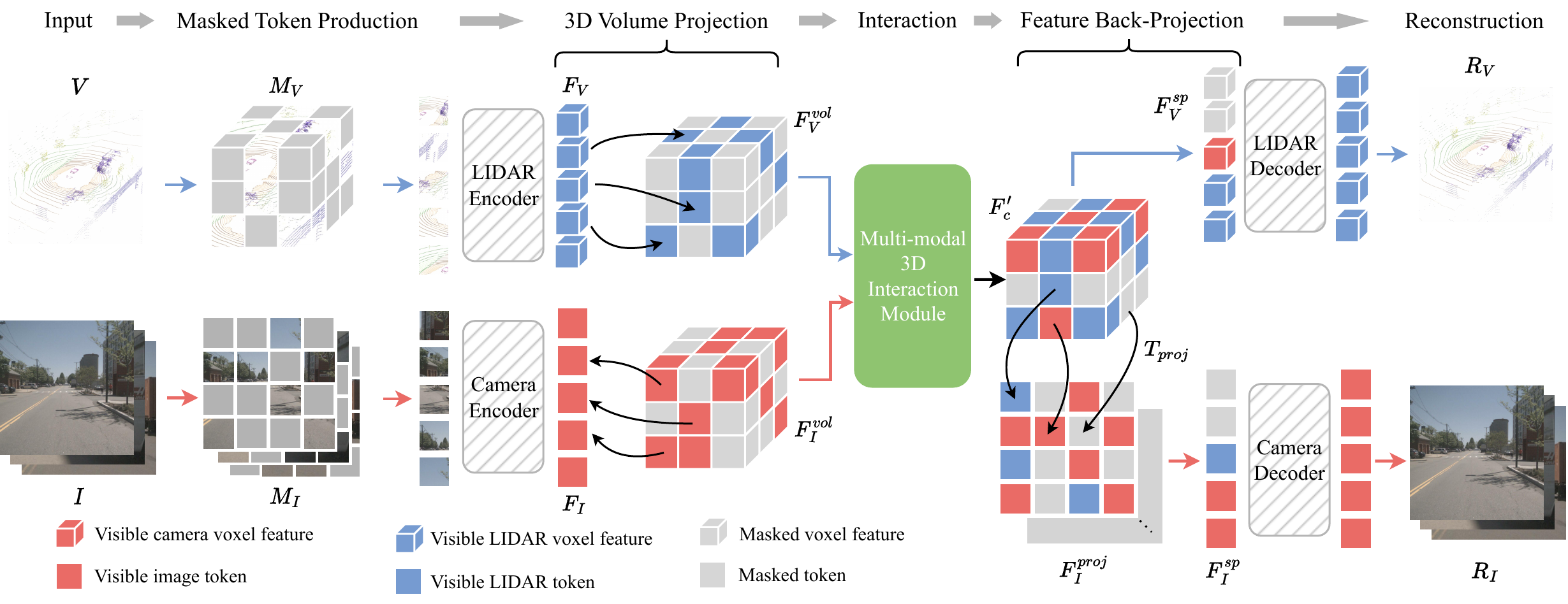}
    \end{center}
    \caption{\textbf{Overview of UniM$^2$AE}. The LiDAR branch voxelize the point cloud, while the camera branch divides multiple images into patches, both subsequently randomly masking their inputs. The tokens from the two branches are individually embedded and then passed through the Token-Volume projection, Multi-modal 3D Interaction Module, Volume-Token projection, and eventually the modality-specific decoder. Ultimately, we reconstruct the original inputs using the fused features.}
    \label{overview}
\end{figure*}

\subsection{Multi-modal Fusion in 3D Perception}

Recently, multi-modal fusion has been well-studied in 3D perception. Proposal-level fusion methods adopt proposals in 3D and project the proposals to images to extract RoI feature~\cite{chen2017multi,nabati2021centerfusion,chen2023futr3d}. Point-level fusion methods usually paint image semantic features onto foreground LiDAR points, which can be classified into input-level decoration ~\cite{vora2020pointpainting, wang2021pointaugmenting, yin2021multimodal}, and feature-level decoration ~\cite{liang2018deep, li2022deepfusion}. However, the camera-to-LiDAR projection is semantically lossy due to the different densities of both modalities. 
Some BEV-based approaches~\cite{chi2023bev} aim to mitigate this problem, but their simple fusion modules fall short in retaining the height information and remain blurry along \textit{z}-axis, which is disadvantageous for accurate image reconstruction.
Accordingly, we design the Multi-modal 3D Interaction Module to effectively fuse the projected 3D volume features.

\section{Proposed Method}

\subsection{Overview Architecture}

As shown in Fig.~\ref{overview}, UniM$^2$AE learns multi-modal representation by masking inputs $(I, V)$ and jointly combine features projected to the 3D volume space $(F^{vol}_V, F^{vol}_I)$ to accomplish the reconstruction. In our proposed pipeline, the point cloud is first embeded into tokens after voxelization and similarly we embed the images with position encoding after dividing the them into non-overlapping patches. 
Following this, tokens from both modalities are randomly masked, producing $(M_I, M_V)$.  Separate transformer-based encoders are then utilized to extract features $(F_I, F_V)$. 

To align features from various modalities with the preservation of semantics and geometrics, $(F_I, F_V)$ are separately projected into the unified 3D volume space, which is extended BEV along the height dimension. Specifically, we build a mapping of each voxel to 3D volume space based on its position in the ego-vehicle coordinates, while for the image tokens, the spatial cross-attention is employed for 2D to 3D conversion. The projected feature $(F^{vol}_V, F^{vol}_I)$ are subsequently passed into the Multi-modal 3D Interaction Module (MMIM), aiming at promoting powerful feature fusion.

Following the cross-modal interaction, we project the fused feature $F^{\prime}_c$ back to the modality-specific token, denoted $F^{sp}_V$ for LiDAR and $F^{proj}_I\in(C, H, W)$ (which is then reshaped to $F^{sp}_I\in(HW, C)$) for camera. The camera decoder and LiDAR decoder are finally used to reconstruct the original inputs.

\subsection{Unified Representation in 3D Volume Space}

Different sensors capture data that, while representing the same scene, often provide distinct descriptions. For instance, camera-derived images emphasize the color palette of the environment within their field of view, whereas point clouds primarily capture object locations. Given these variations, selecting an appropriate representation for fusing features from disparate modalities becomes paramount. Such a representation must preserve the unique attributes of multi-modal information sourced from various sensors.

In pursuit of capturing the full spectrum of object positioning and appearance, the voxel feature in 3D volume space is adopted as the unified representation, depicted in Fig.~\ref{teaser}(b). The 3D volume space uniquely accommodates the height dimension, enabling it to harbor more expansive geometric data and achieve exacting precision in depicting object locations, exemplified by features like elevated traffic signs.
This enriched representation naturally amplifies the accuracy of interactions between objects.
A salient benefit of the 3D volume space is its capacity for direct remapping to original modalities, cementing its position as an optimal latent space for integrating features. 
Due to the intrinsic alignment of images and point clouds within the 3D volume space, the Multi-modal 3D Interaction Module can bolster representations across streams, sidestepping the need for additional alignment mechanisms.
Such alignment streamlines the transition between pre-training and fine-tuning, producing favorable outcomes for subsequent tasks. Additionally, the adaptability of the 3D volume space leaves the door open for its extension to encompass three or even more modalities.

\subsection{Multi-modal Interaction}

\begin{figure}[!htb]
    \begin{center}
        \includegraphics[width=.85\linewidth]{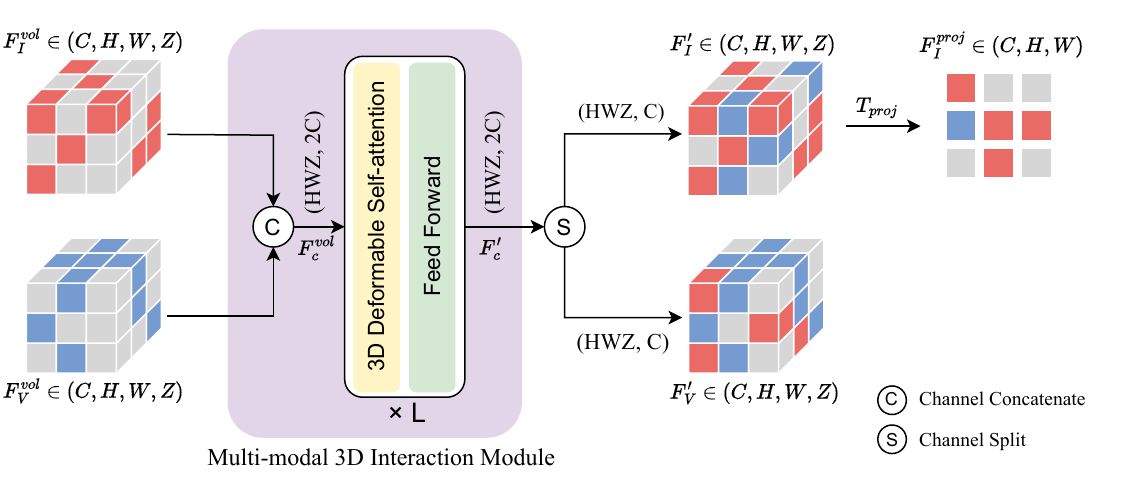}
    \end{center}
    \caption{\textbf{Illustration of our Multi-modal 3D Interaction Module}. We first concatenate the inputs $\left(F_V^{vol}, F_I^{vol}\right)$ and reshape it for the subsequent stacking 3D deformable self-attention blocks. After interaction, we split the output and project them back to feature token. This contributes more generalized and effective feature learning.}
    \label{mmim}
\end{figure}

\subsubsection{Projection to 3D Volume Space}

In the projection of LiDAR to the 3D volume, the voxel embedding is directed to a predefined 3D volume using positions from the ego car coordinate system. This method ensures no geometric distortion is introduced. The resulting feature from this process is denoted as ${F}^{vol}_{V}$.
For the image to 3D volume projection, the 2D-3D Spatial Cross-Attention method is employed. Following prior works \cite{wei2023surroundocc, li2022bevformer}, 3D volume queries for each image is defined as $Q^{vol} \in \mathbb{R}^{C\times H \times W \times Z}$.
The corresponding 3D points are then projected to 2D views using the camera's intrinsic and extrinsic parameters. During this projection, the 3D points associate only with specific views, termed as $\mathcal{V}_{hit}$. The 2D feature is then sampled from the view $\mathcal{V}_{hit}$ at the locations of those projected 3D reference points. The process unfolds as:
\begin{equation}
    F^{vol}_{I} = \frac{1}{\left|\mathcal{V}_{hit}\right|} \sum_{i \in \mathcal{V}_{hit}} \sum_{j = 1}^{N_{ref}} \mathrm{DeformAttn}\left(Q^{vol}, \mathcal{P}(p, i, j), F_{I}^{i}\right)
\end{equation}
where $i$ indexes the camera view, $j$ indexes the 3D reference points, and $N_{ref}$ is the total number of points for each 3D volume query. $F_{I}^{i}$ is the features of the $i$-th camera view. $\mathcal{P}(p, i, j)$ is the project function that gets the $j$-th reference point on the $i$-th view image.

\subsubsection{Multi-modal 3D Interaction Module}

To fuse the projected 3D volume features from the camera($F^{vol}_I\in\mathbb{R}^{C\times H\times W\times Z}$) and the LiDAR($F^{vol}_V\in\mathbb{R}^{C\times H\times W\times Z}$) branches effectively, the Multi-modal 3D Interaction Module (MMIM) is introduced. 
As depicted in Fig.~\ref{mmim}, MMIM comprises $L$ attention blocks, with  $L = 3$  being the default setting.

Given the emphasis on high performance at high resolutions in downstream tasks and the limited scale of token sequences in standard self-attention, deformable self-attention is selected to alleviate computational demands. Each block is composed of deformable self-attention, a feed-forward network, and normalization.
Initially, the concatenation of $F_V^{vol}$ and $F_I^{vol}$ is performed along the channel dimension, reshaping the result to form the query token $F_{c}^{vol} \in \mathbb{R}^{HWZ \times 2C}$.
This token is then inputted into the Multi-modal 3D Interaction Module, an extension of \cite{zhu2020deformable}, to promote effective modal interaction. The interactive process can be described as follows:
\begin{equation}
     F_{c}^{\prime} = \sum_{m = 1}^{M}{W}_{m}\sum_{k = 1}^{K} A_{m k} \cdot {W}_{m}^{\prime} {F_{c}^{vol}}\left({p}^{vol}+\Delta {p}^{vol}_k\right)
\end{equation}
where $m$ indexes the attention head, $k$ indexes the sampled keys, and $K$ is the total sampled key number. $\Delta p_k^{vol}$ and $A_{mk}$ denote the sampling offset and attention weight of the $k$-th sampling point in the $m$-th attention head, respectively. The attention weight $A_{mk}$ lies in the range $\left[0, 1\right]$, normalized by $\sum^K_{k = 1} A_{mk} = 1$. 
At the end, $F_{c}^{\prime}$ is split along the channel dimension to obtain the modality-specific 3D volume features $(F^\prime _{V}, F^\prime _{I})$.

\subsubsection{Projection to Modality-specific Token}

By tapping into the advantages of the 3D volume representation, the fused feature can be conveniently projected onto the 2D image plane and 3D voxel token.
For the LiDAR branch, the process merely involves sampling the features located at the position of the masked voxel token within the ego-vehicle coordinates. Notably, these features have already been enriched by the fusion module with semantics from the camera branch.
Regarding the camera branch, the corresponding 2D coordinate $(u, v)$ can be determined using the projection function $T$. The 2D-plane feature  $F^{proj}_I$ is then obtained by mapping the 3D volume feature in $(x, y, z)$  to the position $(u, v)$.
The projection function $T_{proj}$ is defined as :
\begin{equation}
z\left[\begin{array}{l}
u \\
v \\
1
\end{array}\right] = T_{proj}(P) = K \cdot R_{t} \cdot\left[\begin{array}{l}
x \\
y \\
z \\
1
\end{array}\right]
\end{equation}
where $P \in \mathbb{R}^3$ is the position in 3D volume, $K \in \mathbb{R}^{3 \times 4}$, $R_t \in \mathbb{R}^{4 \times 4}$ are the camera intrinsic and extrinsic matrices.

\subsection{Prediction Target \label{pred}} 

Three distinct reconstruction tasks supervise each modal decoder. A single linear layer is applied to the output of each decoder for each task. The dual-modal reconstruction tasks and their respective loss functions are detailed below.
In alignment with Voxel-MAE \cite{hess2023masked}, the prediction focuses on the number of points within each voxel. Supervision for this reconstruction uses the Chamfer distance, which gauges the disparity between two point sets of different scales.
Let $G_n$ denote the masked LiDAR point cloud partitioned into voxels. 
The Chamfer loss $\mathcal{L}_c$ can be presented as:
\begin{equation}
    \mathcal{L}_c = CD\left(Dec_{V}\left(F^{sp}_V\right), G_n\right)
\end{equation}
where $CD\left(\cdot\right)$ stands for Chamfer distance function \cite{fan2017point}, $Dec_V(\cdot)$ denote voxel decoder and $F^{sp}_V$ represents projected voxel features.

In addition to the aforementioned reconstruction task, there is a prediction to ascertain if a voxel is empty. Supervision for this aspect employs the binary cross entropy loss, denoted as $\mathcal{L}_{occ}$. The cumulative voxel reconstruction loss is thus given as:
\begin{equation}
\mathcal{L}_{voxel} = \mathcal{L}_{c}+\mathcal{L}_{occ}
\end{equation}

For the camera branch, the pixel reconstruction is supervised using the Mean Squared Error (MSE) loss, represented as :
\begin{equation}
\mathcal{L}_{img} = \mathcal{L}_{MSE}\left(Dec_{I}\left(F^{sp}_I\right), G_{I}\right)
\end{equation}
where $G_I$ is the original images in pixel space, $Dec_{I}(\cdot)$ denotes the image decoder and $F^{sp}_I$ represents projected image features.
\section{Experiments}

\subsection{Implementation Details}

\subsubsection{Dataset and Metrics} 
The nuScenes Dataset \cite{caesar2020nuscenes}, a comprehensive autonomous driving dataset, serves as the primary dataset for both pre-training our model and evaluating its performance on multiple downstream tasks. This dataset encompasses 1,000 sequences gathered from Boston and Singapore, with 700 designated for training and 300 split evenly for validation and testing. 
Each sequence, recorded at 10Hz, spans 20 seconds and is annotated at a frequency of 2Hz.
In terms of 3D detection, the principal evaluation metrics employed are mean Average Precision (mAP) and the nuScenes detection score (NDS). 
For the task of BEV map segmentation, the methodology aligns with the dataset's map expansion pack, using Intersection over Union (IoU) as the assessment metric.

\subsubsection{Network Architectures}
The UniM$^2$AE utilizes SST~\cite{fan2022embracing} and Swin-T~\cite{liu2021swin} as the backbones for the LiDAR Encoder and Camera Encoder, respectively. In the Multi-modal 3D Interaction Module, 3 deformable self-attention blocks are stacked, with each attention module comprising 128 input channels and 256 hidden channels. To facilitate the transfer of pre-trained weights for downstream tasks, BEVFusion-SST and TransFusion-L-SST are introduced, with the LiDAR backbone in these architectures being replaced by SST.

\subsubsection{Pre-training}

During this stage, the perception ranges are restricted to [-50m, 50m] for \textit{x}- and \textit{y}-axes, [-5m, 3m] for \textit{z}-axes. Each voxel has dimensions of (0.5m, 0.5m, 4m). 
In terms of input data masking during the training phase, our experiments have determined that a masking ratio of 70\% for the LiDAR branch and 75\% for the camera branch yields optimal results. By default, all the MAE methods are trained with a total of 200 epochs on 8 GPUs, a base learning rate of 2.5e-5. Detailed configurations are reported in the supplemental material.

\begin{table*}[!htb]
\centering
\caption{Data-efficient 3D object detection results of on the nuScenes validation set. Backbones in single-modality and multi-modality are pre-trained using various MAE methods. The model performances are reported using different amounts of fine-tuning data. Random denotes training from scratch. MIM+Voxel-MAE: for the initialization of the model, the image backbone loads the GreenMIM~\cite{huang2022green} pre-trained weights, whereas the point cloud backbone loads the Voxel-MAE~\cite{hess2023masked} pre-trained weight. L and C represent LiDAR and Camera, respectively. *: our re-implementation.}
\resizebox{\textwidth}{!}{%
\begin{tabular}{@{}c|c|c|cc|cccccccccc@{}}
\toprule
Data amount            & Modality             & Initialization & mAP           & NDS           & Car           & Truck         & C.V.          & Bus           & Trailer       & Barrier       & Motor         & Bike          & Ped.          & T.C.          \\ \midrule
\multirow{7}{*}{20\%}  & \multirow{3}{*}{L}   & Random         & 44.3          & 56.3          & 78.8          & 41.1          & \textbf{13.1} & 50.7          & 18.8          & 52.8          & 46.1          & 17.3          & 75.7          & 49.1          \\
                       &                      & Voxel-MAE*     & 48.9          & 59.8          & 80.9          & 47.0          & 12.8          & \textbf{59.0} & 23.6          & 61.9          & 47.8          & 23.5          & 79.9          & 52.4          \\
                       &                      & \textbf{UniM$^2$AE}  & \textbf{50.0} & \textbf{60.0} & \textbf{81.0} & \textbf{47.8} & 12.0          & 57.3          & \textbf{24.0} & \textbf{62.3} & \textbf{51.2} & 26.8          & \textbf{81.4} & \textbf{55.8} \\ \cmidrule(l){2-15} 
                       & \multirow{4}{*}{C+L} & Random         & 51.5          & 50.9          & 84.1          & 47.6          & 13.3          & 49.8          & 27.9          & 65.0          & 53.9          & 26.8          & 78.8          & 67.8          \\
                       &                      & MIM+Voxel-MAE  & 54.3          & 51.2          & 84.3          & 50.8          & 18.9          & 52.3          & 28.9          & 68.4          & 57.4          & 32.1          & 80.3          & 69.2          \\
                       &                      & PiMAE*~\cite{chen2023pimae}  & 52.5          & 52.0          & 84.1          & 48.9          & 17.5          & 50.0          & 28.0          & 66.2          & 51.6          & 26.8          & 80.8          & 70.9          \\
                       &                      & \textbf{UniM$^2$AE}  & \textbf{55.9} & \textbf{52.8} & \textbf{85.8} & \textbf{51.1} & \textbf{19.3} & \textbf{54.2} & \textbf{30.6} & \textbf{69.0} & \textbf{61.1} & \textbf{34.3} & \textbf{83.0} & \textbf{70.8} \\ \midrule
\multirow{7}{*}{40\%}  & \multirow{3}{*}{L}   & Random         & 50.9          & 60.5          & 81.4          & 47.6          & 13.7          & 58.0          & 24.5          & 61.3          & 57.7          & 30.1          & 80.4          & 54.4          \\
                       &                      & Voxel-MAE*     & 52.6          & 62.2          & 82.4          & 49.1          & 15.4          & \textbf{62.2} & \textbf{25.8} & 64.5          & 56.8          & 30.4          & 82.3          & 57.5          \\
                       &                      & \textbf{UniM$^2$AE}  & \textbf{52.9} & \textbf{62.6} & \textbf{82.7} & \textbf{49.2} & \textbf{15.8} & 60.1          & 23.7          & \textbf{65.5} & \textbf{58.4} & \textbf{31.2} & \textbf{83.8} & \textbf{58.9} \\ \cmidrule(l){2-15} 
                       & \multirow{4}{*}{C+L} & Random         & 58.6          & 61.9          & 86.2          & 54.7          & 21.7          & 60.0          & 33.0          & 70.7          & 64.2          & 38.6          & 83.0          & 74.3          \\
                       &                      & MIM+Voxel-MAE  & 60.2          & 63.5          & 86.6          & 56.5          & 22.5          & 64.1          & 33.5          & \textbf{72.2} & 66.1          & 41.9          & 83.8          & 74.8          \\
                       &                      & PiMAE*~\cite{chen2023pimae}  & 61.4          & 64.0          & 86.9          & 57.8          & \textbf{25.2}    & \textbf{64.6} & 36.2          & 70.6          & \textbf{69.8}         & 44.3          & 83.0          & 75.5          \\
                       &                      & \textbf{UniM$^2$AE}  & \textbf{62.0} & \textbf{64.5} & \textbf{87.0} & \textbf{57.8} & 22.8 & 62.7          & \textbf{38.7} & 69.7          & 66.8 & \textbf{50.5} & \textbf{86.0} & \textbf{77.9} \\ \midrule
\multirow{7}{*}{60\%}  & \multirow{3}{*}{L}   & Random         & 51.9          & 61.7          & 82.2          & 49.0          & 15.6          & 61.2          & 24.9          & 62.9          & 56.3          & 32.1          & 81.6          & 53.1          \\
                       &                      & Voxel-MAE*     & 54.2          & 63.5          & 83.0          & \textbf{51.1} & 16.3          & 62.0          & \textbf{27.5} & 64.9          & 61.2          & 34.7          & 82.9          & 58.0          \\
                       &                      & \textbf{UniM$^2$AE}  & \textbf{54.7} & \textbf{63.8} & \textbf{83.1} & 51.0          & \textbf{17.3} & \textbf{62.5} & 26.9          & \textbf{65.1} & \textbf{62.2} & \textbf{35.7} & \textbf{83.4} & \textbf{59.9} \\ \cmidrule(l){2-15} 
                       & \multirow{4}{*}{C+L} & Random         & 61.6          & 65.2          & 87.3          & 58.5          & 23.9 & 65.2          & 35.8          & 71.9 & 67.8          & 46.7          & 85.7          & 77.0          \\
                       &                      & MIM+Voxel-MAE  & 62.1          & 65.7          & 87.2          & 56.7          & 23.0          & 65.4          & 37.0          & 71.7          & \textbf{70.6} & 47.3          & 85.6          & 76.7          \\
                       &                      & PiMAE*~\cite{chen2023pimae}  & 62.3          & 65.5          & 87.3          & 58.8          & \textbf{24.9}    & 61.6          & \textbf{38.7} & \textbf{72.7}    & 68.5       & 45.7          & 86.0          & \textbf{79.1}          \\
                       &                      & \textbf{UniM$^2$AE}  & \textbf{62.4} & \textbf{66.1} & \textbf{87.7} & \textbf{59.7} & 23.8 & \textbf{67.6} & 37.0 & 70.5          & 68.4          & \textbf{48.9} & \textbf{86.6} & 77.8 \\ \midrule
\multirow{7}{*}{80\%}  & \multirow{3}{*}{L}   & Random         & 52.7          & 62.5          & 82.3          & 49.6          & 16.0          & 63.3          & 25.8          & 60.7          & 58.6          & 31.6          & 82.0          & 56.7          \\
                       &                      & Voxel-MAE*     & 55.1          & 64.2          & 83.4          & 51.7          & \textbf{18.8} & 64.0          & 28.7          & 63.8          & 62.2          & 35.1          & 84.3          & 58.7          \\
                       &                      & \textbf{UniM$^2$AE}  & \textbf{55.6} & \textbf{64.6} & \textbf{83.4} & \textbf{52.9} & 18.2          & \textbf{64.2} & \textbf{29.4} & \textbf{64.7} & \textbf{63.1} & \textbf{36.1} & \textbf{84.5} & \textbf{58.8} \\ \cmidrule(l){2-15} 
                       & \multirow{4}{*}{C+L} & Random         & 62.5          & 66.1          & 87.1          & 57.6          & 24.0          & 66.4          & 38.1          & 71.1          & 68.4          & 48.8          & 86.2          & 77.6          \\
                       &                      & MIM+Voxel-MAE  & 63.0          & 66.4          & 87.6          & 59.6          & 24.1          & 66.1          & 38.0          & 71.3          & 70.2          & 48.8          & 86.5          & 78.1          \\
                       &                      & PiMAE*~\cite{chen2023pimae}  & 63.7          & 66.6          & 87.0          & \textbf{61.9}    & \textbf{25.3} & \textbf{71.2}          & 38.6          & 70.3          & \textbf{73.9}          & 49.3          & 84.6          & 74.7          \\
                       &                      & \textbf{UniM$^2$AE}  & \textbf{63.9} & \textbf{67.1} & \textbf{87.7} & 59.6 & 24.9 & 69.2 & \textbf{39.8} & \textbf{71.3} & 71.0 & \textbf{50.2} & \textbf{86.8} & \textbf{78.7} \\ \midrule
\multirow{7}{*}{100\%} & \multirow{3}{*}{L}   & Random         & 53.6          & 63.0          & 82.3          & 49.8          & 16.7          & 64.0          & 26.2          & 60.9          & 61.7          & 33.0          & 82.2          & 58.9          \\
                       &                      & Voxel-MAE*     & 55.3          & 64.1          & 83.2          & 51.2          & 16.8          & \textbf{64.6} & 28.3          & 65.0          & 61.8          & 39.6          & 83.6          & 58.9          \\
                       &                      & \textbf{UniM$^2$AE}  & \textbf{55.8} & \textbf{64.6} & \textbf{83.3} & \textbf{51.3} & \textbf{17.6}  & 63.7          & \textbf{28.6} & \textbf{65.4} & \textbf{62.8} & \textbf{40.8} & \textbf{83.9} & \textbf{60.3} \\ \cmidrule(l){2-15} 
                       & \multirow{4}{*}{C+L} & Random         & 63.6          & 67.4          & 87.7          & 58.0          & 26.6 & 67.8          & 38.4          & \textbf{72.6} & 71.8          & 47.6          & 87.0          & 78.9          \\
                       &                      & MIM+Voxel-MAE  & 63.7          & 67.7          & 87.6          & 58.3 & 25.3          & 67.1          & 38.8          & 70.8          & 71.7          & \textbf{51.5} & 86.7          & 78.9          \\
                       &                      & PiMAE*~\cite{chen2023pimae}  & 63.9          & 67.9          & 87.3          & \textbf{58.7} & \textbf{27.0}   & 67.6          & 38.7          & 69.7          & 71.4          & \textbf{55.4} & 86.8          & 76.6          \\
                       &                      & \textbf{UniM$^2$AE}  & \textbf{64.3} & \textbf{68.1} & \textbf{87.9} & 57.8          & 24.3          & \textbf{68.6} & \textbf{42.2} & 71.6          & \textbf{72.5} & 51.0          & \textbf{87.2} & \textbf{79.5} \\ \bottomrule
\end{tabular}%
}
\label{data_effi}
\end{table*}

\subsubsection{Fine-tuning}
Utilizing the encoders from UniM$^2$AE for both camera and LiDAR, the process then involves fine-tuning and assessing the capabilities of the features learned on tasks that are both single-modal and multi-modal in nature.
For tasks that are solely single-modal, one of the pre-trained encoder serves as the feature extraction mechanism. 
When it comes to multi-modal tasks, which include 3D object detection and BEV map segmentation, both the LiDAR encoder and the camera branch's encoder are capitalized upon as the feature extractors pertinent to these downstream tasks. It's pivotal to note that while the decoder plays a role during the pre-training phase, it's omitted during the fine-tuning process.
A comparison of the pre-trained feature extractors with a variety of baselines across different tasks was conducted, ensuring the experimental setup remained consistent. 
Notably, when integrated into fusion-based methodologies, the pre-trained Multi-modal 3D Interaction Module showcases competitive performance results. Detailed architectures and configurations are in the supplemental material.

\subsection{Data Efficiency}

The primary motivation behind employing MAE is to minimize the dependency on annotated data without compromising the efficiency and performance of the model. In assessing the representation derived from the pre-training with UniM$^2$AE, experiments were conducted on datasets of varying sizes, utilizing different proportions of the labeled data. Notably, for training both the single-modal and multi-modal 3D Object Detection models, fractions of the annotated dataset, namely $\{20\%, 40\%, 60\%, 80\%, 100\%\}$, are used.

\begin{table*}[!thb]
\centering
\caption{Performances of 3D object detection on the nuScenes validation split. $^{\dag}$: Fine-tuned with the pre-trained MMIM. *: Our re-implementation with SST backbone.}
\resizebox{\textwidth}{!}{%
\begin{tabu}{lccccccccc}
\toprule
Method            & Modality & Voxel Size(m)           & NDS↑ & mAP↑ & mATE↓ & mASE↓         & mAOE↓ & mAVE↓ & mAAE↓         \\ \midrule
CenterPoint~\cite{yin2021center} & L        & {[}0.075, 0.075, 0.2{]} & 66.8 & 59.6 & 29.2  & 25.5          & 30.2  & 25.9  & 19.4       \\
LargeKernel3D~\cite{chen2023largekernel3d} & L        & {[}0.075, 0.075, 0.2{]} & 69.1 & 63.9 & 28.6  & 25.0          & 35.1  & 21.1  & 18.7  \\
TransFusion-L~\cite{bai2022transfusion} & L        & {[}0.075, 0.075, 0.2{]} & 70.1 & 65.4 & -     & -             & -     & -     & -    \\ 
\tabucline [1pt on 1.5pt off 2pt]{1-10}
TransFusion-L-SST & L        & {[}0.15, 0.15, 8{]}     & 69.9 & 65.0 & 28.0  & 25.3          & 30.1  & 24.1  & 19.0          \\
TransFusion-L-SST+\textbf{UniM$^2$AE-L} & L   & {[}0.15, 0.15, 8{]} & \textbf{70.4} & \textbf{65.7} & \textbf{28.0} & \textbf{25.2} & \textbf{29.5} & \textbf{23.5} & \textbf{18.6} \\ \hline
FUTR3D~\cite{chen2023futr3d} & C+L      & {[}0.075, 0.075, 0.2{]} & 68.3 & 64.5 & -     & -             & -     & -     & -             \\
Focals Conv~\cite{chen2022focal} & C+L      & {[}0.075, 0.075, 0.2{]} & 69.2 & 64.0 & 33.2  & 25.4          & 27.8  & 26.8  & 19.3          \\
MVP~\cite{yin2021multimodal} & C+L      & {[}0.075, 0.075, 0.2{]} & 70.7 & 67.0 & 28.9  & 25.1 & 28.1  & 27.0  & 18.9          \\
TransFusion~\cite{bai2022transfusion} & C+L      & {[}0.075, 0.075, 0.2{]} & 71.2 & 67.3 & 27.2  & 25.2          & 27.4  & 25.4  & 19.0          \\
MSMDFusion~\cite{jiao2023msmdfusion} & C+L      & {[}0.075, 0.075, 0.2{]} & 72.1 & 69.3 & -     & -             & -     & -     & -             \\
BEVFusion~\cite{liu2023bevfusion} & C+L      & {[}0.075, 0.075, 0.2{]} & 71.4 & 68.5 & 28.7  & 25.4          & 30.4  & 25.6  & \textbf{18.7} \\
\tabucline [1pt on 1.5pt off 2pt]{1-10}
BEVFusion-SST     & C+L      & {[}0.15, 0.15, 8{]}     & 71.5 & 68.2 & 27.8  & 25.3          & 30.2  & 23.6  & 18.9          \\
BEVFusion-SST+UniM$^2$AE  & C+L & {[}0.15, 0.15, 8{]} & 71.9          & 68.4          & 27.2          & 25.2          & 28.8          & 23.2 & \textbf{18.7} \\
BEVFusion-SST+\textbf{UniM$^2$AE$^{\dag}$}   & C+L & {[}0.15, 0.15, 4{]} & 72.7 & 69.7 & \textbf{26.9} & 25.2    & 27.3 & 23.2 & 18.9 \\ 
\tabucline [1pt on 1.5pt off 2pt]{1-10}
FocalFormer3D~\cite{chen2023focalformer3d} & C+L & {[}0.075, 0.075, 0.2{]} & 73.1 & 70.5 & - & - & - & - & - \\
FocalFormer3D*+\textbf{UniM$^2$AE$^{\dag}$}   & C+L & {[}0.15, 0.15, 4{]} & \textbf{73.8} & \textbf{71.1} & \textbf{26.9} & \textbf{25.0} & \textbf{26.7} & \textbf{19.6} & 18.9 
\\ \bottomrule
\end{tabu}%
}
\label{nuscenes-val}
\end{table*}

In the realm of single-modal 3D self-supervised techniques, UniM$^2$AE is juxtaposed against Voxel-MAE~\cite{hess2023masked} using an anchor-based detector. Following the parameters set by Voxel-MAE, the detector undergoes training for 288 epochs with a batch size of 4 and an initial learning rate pegged at 1e-5.
On the other hand, for multi-modal strategies, evaluations are conducted on a fusion-based detector~\cite{liu2023bevfusion} equipped with a TransFusion head~\cite{bai2022transfusion}. As per current understanding, this is a pioneering attempt at implementing multi-modal MAE in autonomous driving. For the sake of a comparative analysis, a combination of the pre-trained Swin-T from GreenMIM~\cite{huang2022green} and SST from Voxel-MAE~\cite{hess2023masked} is utilized.

According to the result in Table \ref{data_effi}, the proposed UniM$^2$AE presents a substantial enhancement to the detector, exhibiting an improvement of $4.4/1.9$ mAP/NDS over random initialization and $1.6/1.6$ mAP/NDS compared to the basic amalgamation of GreenMIM and Voxel-MAE when trained on just $20\%$ of the labeled data. Impressively, even when utilizing the entirety of the labeled dataset, UniM$^2$AE continues to outperform, highlighting its superior ability to integrate multi-modal features in the unified 3D volume space during the pre-training phase.
Moreover, it's noteworthy to mention that while UniM$^2$AE isn't specifically tailored for a LiDAR-only detector, it still yields competitive outcomes across varying proportions of labeled data. This underscores the capability of UniM$^2$AE to derive more insightful representations.

\begin{table*}[!htb]
\centering
\caption{Performances of the BEV map segmentation on the nuScenes validation split. $^{\dag}$: Fine-tuned with the pre-trained MMIM. *: re-implemented by training from scratch.}
\resizebox{\textwidth}{!}{%
\begin{tabu}{lcccccccccc}
\toprule
Method              & Modality & Drivable      & Ped. Cross.   & Walkway       & Stop Line     & Carpark       & Divider       & mIoU          \\ \midrule
CVT~\cite{zhou2022cross} & C        & 74.3          & 36.8          & 39.9          & 25.8          & 35.0          & 29.4          & 40.2     \\
OFT~\cite{roddick2018orthographic} & C        & 74.0          & 35.3          & 45.9          & 27.5          & 35.9          & 33.9          & 42.1          \\
LSS~\cite{philion2020lift} & C        & 75.4          & 38.8          & 46.3          & 30.3          & 39.1          & 36.5          & 44.4   \\
M$^2$BEV~\cite{xie2022m} & C        & 77.2          & -             & -             & -             & -             & 40.5          & -        \\ 
\tabucline [1pt on 1.5pt off 2pt]{1-9}
BEVFusion*~\cite{liu2023bevfusion} & C        & 78.2          & 48.0          & 53.5          & 40.4          & 45.3          & 41.7          & 51.2          \\
BEVFusion*+\textbf{UniM$^2$AE-C} & C  & \textbf{79.5} & \textbf{50.5} & \textbf{54.9} & \textbf{42.4} & \textbf{47.3} & \textbf{42.9} & \textbf{52.9}       \\ \midrule
MVP~\cite{yin2021multimodal} & C+L      & 76.1          & 48.7          & 57.0          & 36.9          & 33.0          & 42.2          & 49.0          \\
PointPainting~\cite{vora2020pointpainting} & C+L      & 75.9          & 48.5          & 57.7          & 36.9          & 34.5          & 41.9          & 49.1          \\
BEVFusion~\cite{liu2023bevfusion} & C+L      & 85.5          & 60.5          & 67.6          & 52.0          & 57.0          & 53.7          & 62.7          \\
X-Align~\cite{borse2023x} & C+L      & 86.8          & 65.2          & 70.0          & 58.3          & 57.1          & 58.2          & 65.7          \\
\tabucline [1pt on 1.5pt off 2pt]{1-9}
BEVFusion-SST & C+L      & 84.9          & 59.2          & 66.3          & 48.7          & 56.0          & 52.7          & 61.3          \\
BEVFusion-SST+UniM$^2$AE & C+L      & 85.1          & 59.7          & 66.6          & 48.7          & 56.0          & 52.6          & 61.4          \\
BEVFusion-SST+\textbf{UniM$^2$AE$^{\dag}$} & C+L      & \textbf{88.7} & \textbf{67.4} & \textbf{72.9} & \textbf{59.0} & \textbf{59.0} & \textbf{59.7} & \textbf{67.8} \\ \bottomrule
\end{tabu}%
}
\label{nuscenes-seg}
\end{table*}

PiMAE~\cite{chen2023pimae}, as the first framework to explore multi-modal MAE in indoor scenarios, projects point clouds onto the image plane aligned and masked with the image, followed by concatenation of dual modality tokens along the sequence dimension to pre-train the ViT backbone.
We compare the UniM$^2$AE to the previous SOTA multi-modal pre-training method \ie PiMAE, on the nuScenes benchmark~\cite{caesar2020nuscenes}.
As shown in Table~\ref{data_effi}, when trained with $20\%$ labeled data, PiMAE outperforms training from scratch with a marginal increase of $1.1/1.0$ NDS/mAP due to the loss of depth information in aligning point clouds with images in the pixel coordinates. 
Conversely, UniM$^2$AE, aligning in the unified 3D volume space, substantially reduces information loss, boosting downstream model by $1.9/4.4$ NDS/mAP, respectively. 
The utilization of MMIM also accelerates pre-training convergence, which is also the key to improving downstream task performance.

\subsection{Comparison on Downstream Tasks}

\subsubsection{3D Object Detection}

To demonstrate the capability of the learned representation, we fine-tune various pre-trained detectors on the nuScenes dataset. As shown in Table \ref{nuscenes-val}, our UniM$^2$AE substranitally improves both LiDAR-only and fusion-based detection models.
Compared to TransFusion-L-SST, the UniM$^2$AE-L registers a 0.5/0.7 NDS/mAP enhancement on the nuScenes validation subset. In the multi-modality realm, the UniM$^2$AE elevates the outcomes of BEVFusion-SST by 1.2/1.5 NDS/mAP and improves FocalFormer3D~\cite{chen2023focalformer3d} by 0.7/0.6 NDS/mAP when MMIM is pre-trained.
Of note is that superior results are attained even when a larger voxel size is employed. This is particularly significant given that Transformer-centric strategies (\eg SST~\cite{fan2022embracing}) generally trail CNN-centric methodologies (\eg VoxelNet~\cite{yan2018second}).

\subsubsection{BEV Map Segmentation}

Table \ref{nuscenes-seg} presents our BEV map segmentation results on the nuScenes dataset based on BEVFusion~\cite{liu2023bevfusion}. For the camera modality, we outperform the results of training from scratch by 1.7 mAP. In the multi-modality setting, the UniM$^2$AE boosts the BEVFusion-SST by 6.5 mAP with pre-trained MMIM and achieve 2.1 improvement over state-of-the-art methods X-Align~\cite{borse2023x}, indicating the effectiveness and strong generalization of our UniM$^2$AE.

\begin{table*}[!htb]
\centering
\caption{Comparison of the input modality and the interaction space.}
\resizebox{.75\linewidth}{!}{%
\begin{tabular}{cccccccccc}
\toprule
\multicolumn{2}{c}{Modality} &  & \multicolumn{2}{c}{Interaction Space} &  & \multirow{2}{*}{\makebox[0.09\textwidth][c]{mAP}} & \multirow{2}{*}{} & \multirow{2}{*}{\makebox[0.07\textwidth][c]{NDS}} \\ \cline{1-2} \cline{4-5}
\makebox[0.13\textwidth][c]{Camera}    & \makebox[0.12\textwidth][c]{LiDAR} &  & \makebox[0.11\textwidth][c]{BEV} & \makebox[0.14\textwidth][c]{3D Volume}  &  &      &  &      \\ \midrule
           &            &  &            &            &  & 59.0 &  & 61.8 \\
\checkmark &            &  &            &            &  & 59.7 &  & 62.6 \\
           & \checkmark &  &            &            &  & 60.1 &  & 62.6 \\ \midrule
\checkmark & \checkmark &  &            &            &  & 60.7 &  & 63.1 \\
\checkmark & \checkmark &  & \checkmark &            &  & 62.0 &  & 64.3 \\
\checkmark & \checkmark &  &            & \checkmark &  & \textbf{62.8} &  & \textbf{65.2} \\ \bottomrule
\end{tabular}%
}
\label{Tab: module-ab}
\end{table*}

\begin{table}[!htb]
    \caption{Ablation experiments to validate our settings. Detection results are reported on the nuScenes validation split} 
    \label{Tab: ablation study}
    \begin{subtable}{.5\linewidth}
    \centering
    \caption{Impact of the masking ratio.}
    \resizebox{.9\linewidth}{!}{%
    \begin{tabular}{ccccccccccc}
    \toprule
     & \multicolumn{4}{c}{Masking Ratio} &  &  & \multirow{2}{*}{\makebox[0.2\textwidth][c]{mAP}} &  & \multirow{2}{*}{\makebox[0.15\textwidth][c]{NDS}} &  \\ \cline{1-6}
     & \makebox[0.3\textwidth][c]{Camera}  &     &     & \makebox[0.3\textwidth][c]{LiDAR}    &  &  &                      &  &                      &  \\ \midrule
     & 60\%       &     &     & 60\%     &  &  &  63.6                &  &  66.5                &  \\
     & 70\%       &     &     & 70\%     &  &  &  64.2                &  &  \textbf{67.3}       &  \\
     & 75\%       &     &     & 70\%     &  &  &  \textbf{64.5}       &  &  \textbf{67.3}       &  \\
     & 80\%       &     &     & 80\%     &  &  &  63.3                &  &  66.4                &  \\ \bottomrule
    \end{tabular}%
    }
    \label{masking_ratio}
    \end{subtable}%
    \begin{subtable}{.5\linewidth}
      \centering
        \caption{Impact of the number of layers along \textit{z}-axis.}
        \resizebox{\linewidth}{!} {
        \begin{tabular}{ccccc}
            \toprule
            \makebox[0.23\textwidth][c]{Z-layers} & \makebox[0.17\textwidth][c]{NDS} & \makebox[0.17\textwidth][c]{mAP} & \makebox[0.24\textwidth][c]{FLOPs} & \makebox[0.32\textwidth][c]{GPU memory}\\
            \midrule
             1 & 56.6 & 54.9 & 77G & 5.84G        \\
             2 & 58.3 & \textbf{55.0} & 79G & 6.29G \\
             4 & 58.2 & \textbf{55.0} & 88G & 7.16G \\
             8 & \textbf{58.4} & \textbf{55.0} & 105G & 8.80G \\
            \bottomrule
        \end{tabular}
        }
    \label{Tab: z-axis}
    \end{subtable}%
\end{table}

\subsection{Ablation Study}

\subsubsection{Multi-modal Pre-training}

To underscore the importance of the Multi-modal 3D Interaction Module (MMIM) in the 3D volume space for dual modalities, ablation studies were performed with single modal input and were compared with other interaction techniques. 
As presented in Table \ref{Tab: module-ab}, the $1^\text{st}$ row shows the result of training from scratch. The $2^\text{nd}$ and $3^\text{rd}$ rows show fine-tuning results with parameters pre-trained on a single modality. The $4^\text{th}$ row displays outcomes using the two uni-modal pre-training parameters. The $5^\text{th}$ and $6^\text{th}$ rows present fine-tuning results with weights pre-trained with different feature spaces for integration.
Results reveal that by utilizing MMIM to integrate features from two branches within a unified 3D volume space, the UniM$^2$AE model achieves a remarkable enhancement in performance.
Specifically, there's a 3.4 NDS improvement for training from scratch, 2.6 NDS for LiDAR-only and camera-only pre-training, and 2.1 NDS when simply merging the two during the initialization of downstream detectors.
Additionally, a noticeable decline in performance becomes evident when substituting the 3D volume space with BEV, as showcased in the concluding rows of Table \ref{Tab: module-ab}. 
This drop is likely attributed to features mapped onto BEV losing essential geometric and semantic details, especially along the height axis, resulting in a less accurate representation of an object's true height and spatial positioning.
These findings conclusively highlight the crucial role of concurrently integrating camera and LiDAR features within the unified 3D volume space and further validate the effectiveness of the MMIM.

\subsubsection{Masking Ratio}

In Table \ref{masking_ratio}, an examination of the effects of the masking ratio reveals that optimal performance is achieved with a high masking ratio (70\% and 75\%). This not only offers advantages in GPU memory savings but also ensures commendable performance. On the other hand, if the masking ratio is set too low or excessively high, there is a notable decline in performance, akin to the results observed with the single-modal MAE.

\subsubsection{Number of Layers along \textit{z}-axis}

As demonstrated in Table \ref{Tab: z-axis}, an increase in the number of layers beyond two does not uniformly enhance performance, yet it invariably escalates the computational cost. 
This observation can be intuitively understood by considering the typical consistency of road conditions and the specific altitudes where objects are located within a scene(\eg street lights are found above, whereas cars are on the road). Such spatial distribution suggests that a high resolution along the \textit{z}-axis may not always contribute to a meaningful improvement, especially when the additional computational burden is taken into account. Therefore, to strike an optimal balance between computational efficiency and model effectiveness, we opted for two layers along \textit{z}-axis.

\section{Conclusion}

The disparity in the multi-modal integration of MAE methods for practical driving sensors is identified. With the introduction of UniM$^2$AE, a multi-modal self-supervised model is brought forward that adeptly marries image semantics to LiDAR geometries. Two principal innovations define this approach: firstly, the fusion of dual-modal attributes into an augmented 3D volume, which incorporates the height dimension absent in BEV; and secondly, the deployment of the Multi-modal 3D Interaction Module that guarantees proficient cross-modal communications. Benchmarks conducted on the nuScenes Dataset reveal substantial enhancements in 3D object detection by 1.2/1.5 NDS/mAP and in BEV map segmentation by 6.5 mAP, reinforcing the potential of UniM$^2$AE in advancing autonomous driving perception.

\subsubsection{Limitations}
UniM$^2$AE uses a random masking strategy for sensor data from different modalities, which fails to consider their interconnections. Employing simultaneous or complementary masking across multi-modal inputs may enhance feature learning. 
Moreover, dataset in autonomous driving exhibits significant temporal continuity, with adjacent frames frequently resembling each other. This aspect is omitted in our UniM$^2$AE, leading to pre-training on redundant data, which diminishes pre-training efficiency.

\subsubsection{Acknowledgments}
This work was partially supported by National Key RD Program of China under Grant No. 2021ZD0112100.
Guanglei Yang is supported by the Postdoctoral Fellowship Program of CPSF under Grant Number GZC20233458. 
Tao Luo is supported by the A*STAR Funding under Grant Number C230917003.

%
%
\bibliographystyle{splncs04}
\bibliography{main}
\end{document}